\documentclass[conference]{IEEEtran}
\usepackage{balance}       % to better equalize the last page
\usepackage{booktabs}
\usepackage{enumitem}
\usepackage{color}
\usepackage{xspace}
\usepackage{cite}
\usepackage{amsmath,amssymb,amsfonts}
\usepackage{algorithmic}
\usepackage{graphicx}
\usepackage{textcomp}
\usepackage{xcolor}
\usepackage[draft, bookmarks=false]{hyperref}

\def\BibTeX{{\rm B\kern-.05em{\sc i\kern-.025em b}\kern-.08em
    T\kern-.1667em\lower.7ex\hbox{E}\kern-.125emX}}

\newcommand{\hidecomment}[1]{}

\begin{document}

\title{Valuing Player Actions in Counter-Strike: Global Offensive\\
}

\author{\IEEEauthorblockN{Peter Xenopoulos}
\IEEEauthorblockA{New York University\\
New York, NY\\
\href{mailto:xenopoulos@nyu.edu}{xenopoulos@nyu.edu}}
\and
\IEEEauthorblockN{Harish Doraiswamy}
\IEEEauthorblockA{New York University\\
New York, NY\\
\href{mailto:harishd@nyu.edu}{harishd@nyu.edu}}
\and
\IEEEauthorblockN{Claudio Silva}
\IEEEauthorblockA{New York University\\
New York, NY\\
\href{mailto:csilva@nyu.edu}{csilva@nyu.edu}}}
\IEEEoverridecommandlockouts
\IEEEpubid{\makebox[\columnwidth]{978-1-7281-6251-5/20/\$31.00 \textcopyright 2020 IEEE \hfill} \hspace{\columnsep}\makebox[\columnwidth]{ }}

\maketitle
\IEEEpubidadjcol

\begin{abstract}
Esports, despite its expanding interest, lacks fundamental sports analytics resources such as accessible data or proven and reproducible analytical frameworks. Even Counter-Strike: Global Offensive (CSGO), the second most popular esport, suffers from these problems. Thus, quantitative evaluation of CSGO players, a task important to teams, media, bettors and fans, is difficult. To address this, we introduce (1)~a data model for CSGO with an open-source implementation; (2)~a graph distance measure for defining distances in CSGO; and (3)~a context-aware framework to value players' actions based on changes in their team's chances of winning. Using over 70 million in-game CSGO events, we demonstrate our framework's consistency and independence compared to existing valuation frameworks. We also provide use cases demonstrating high-impact play identification and uncertainty estimation.
\end{abstract}


\section{Introduction}
Esports, or professional video gaming, is one of the fastest growing sports in the world. The advent of video streaming has allowed esports to garner viewership as popular sports events like the FIFA World Cup or the College Football National Championship. Accordingly, there has been increased interest in esports from investors looking to create teams, gamblers seeking to wager, and media conglomerates pursuing new, captivated audiences~\cite{keiper2017no}. Yet, esports has attracted limited sports analytics work, unlike traditional sports such as baseball, basketball or soccer.

A common task in sports analytics is to derive quantitative valuations of players. Valuing players is crucial for teams, bettors, media and fans alike. For example, teams may use valuation data to drive efficient player acquisition, bettors may use quantitative metrics to find profitable bets, and media organizations may create data-derived player rankings for curious fans. However, literature on valuing esports players is sparse. Existing frameworks rely on simple, non-contextual statistics that count player outcomes, such as kills, deaths and assists. A downside to this approach is that players are currently being valued as if they are each facing the same situations. However, this is seldom the case, since player outcomes are heavily dependent on context, as some game situations are inherently harder or easier than others. Furthermore, existing frameworks can be hard to reproduce and lack measures of uncertainty.

The key factor affecting the development of involved analytics in esports is the lack of easily accessible and clean esports data.
Improved data capture, along with publicly accessible data, has traditionally enabled analytics to grow in a sport~\cite{assunccao2018sports}. 
Since esports are played entirely virtually, one would expect data capture to be straightforward. However, the tools for data acquisition are surprisingly limited and undocumented. Furthermore, esports data is often stored in esoteric formats that are riddled with inconsistencies and lacks a commonly accepted data model~\cite{bednarek2017data}. These aforementioned issues have hampered the availability of easy to use public esports data, severely limiting the analytics efforts.

In this paper, we provide the following contributions towards analyzing esports data and valuing players. We first define a data model for Counter-Strike: Global Offensive (CSGO), one of the most popular first person shooter games in the world. 
% \claudio{the contributions paragraph does not seem to be the best place to list the related work} Following advances in baseball~\cite{baumer2015openwar}, American football~\cite{yurko2019nflwar} and soccer~\cite{decroos2019actions} that provide implementations of their data model, we provide an open-source library for analyzing CSGO matches using existing data. 
% We also make available the implementation of our data model as an open-source library~\footnote{\href{https://github.com/pnxenopoulos/csgo}{\texttt{csgo} GitHub (https://github.com/pnxenopoulos/csgo)}}.
We will also make available the implementation of our data model as an open-source library~\footnote{\href{https://github.com/pnxenopoulos/csgo}{\texttt{csgo} GitHub (https://github.com/pnxenopoulos/csgo)}}. Next, we introduce a graph-based measure to define distances in CSGO, which we then use to derive spatial features from the data. Finally, we outline a reproducible, context-aware player valuation framework based on how players change their team's chance of winning. Like similar frameworks in other sports, we call our framework Win Probability Added (WPA). We demonstrate the effectiveness of our framework, using over 70 million CSGO events, with a variety of use cases. We find that WPA is both a consistent and unique dimension to value CSGO players, compared to existing frameworks.

The rest of the paper is structured as follows. In Section~\ref{sec:related-work}, we provide a literature review of quantitative player valuation in sports. In Section~\ref{sec:csgo-data}, we define our data model and our graph-based measure for describing distances in CSGO. In Section~\ref{sec:valuation-framework}, we introduce our context-aware player valuation framework and setup for predicting win probability. In Section~\ref{sec:results}, we describe our results and provide insight into our model's calibration. In Section~\ref{sec:discussion}, we compare our framework with existing ones and demonstrate use cases such as identifying high-impact plays and estimating uncertainty. Finally, we conclude the paper in Section~\ref{sec:conclusion}.
\section{Related Work}
\label{sec:related-work}

One of the fundamental objectives of sports analytics is to quantitatively evaluate players. Although major sports, such as soccer and American football, have extensive player valuation literature, esports lags behind. A common player valuation approach is to assess players based on the cumulative value of their actions, where actions are valued by how they change their team's chances of winning or scoring. This approach was employed effectively in soccer, where Decroos~et~al.~\cite{decroos2019actions} presented a machine learning approach to analyze how player actions change their team's chance of scoring or conceding. Their approach estimated the probability of scoring from the most recent sequences of game events. As player actions transition the game from one discrete state to the next, players are valued by how they change their team's chance of scoring or conceding. For American football, Yurko~et~al.~\cite{yurko2019nflwar} not only introduced public NFL play-by-play data, but also a regression based framework to value players based on how their plays change a team's win probability or expected score. Specifically, they estimated expected points and win probability at each play, and valued players by how they changed their team's expected score. 

Central to the aforementioned player valuation frameworks is the idea of estimating win probability or points scored from a given game state. Early attempts used methods such as random forests and ridge regression, applied on discrete play level data, to estimate win probability or points, such as work by Lock~et~al.~\cite{lock2014using} for football and Macdonald~\cite{macdonald2012expected} for ice hockey. As sports data has become more granular, we have seen more continuous time and spatiotemporal models. Cervone~et~al.~\cite{cervone2016multiresolution} utilized Markov chains to estimate the expected points of a basketball possession in real time. More recently, we have seen more neural inspired approaches. Yurko~et~al.~\cite{yurko2019going} designed a framework to estimate American football play values in near-continuous time using player tracking data. Their approach used a long short-term memory recurrent neural network to estimate the expected gain in yards at any point of a play, conditioned on the spatial characteristics of players. Fernandez~et~al.~\cite{fernandez2019decomposing} developed a deep learning approach to estimate the expected goals of a soccer possession using player tracking data to estimate the likelihood of scoring or conceding. Similarly, Sicilia~et~al.~\cite{sicilia2019deephoops} presented a deep learning framework to estimate the value of micro-actions in basketball. Finally, Liu~et~al.~\cite{liu2018deep} used a deep reinforcement learning approach to value ice hockey players as the aggregate of the value of their actions.

Specific to esports, Yang~et~al.~\cite{yang2016real} first estimated win probabilities in Defense of the Ancients 2 (DOTA2), a popular esport game, using logistic regression, and considered both real-time and pre-match factors. Hodge~et~al.~\cite{hodge2019win} attained similar performance on DOTA2 data and suggested that ensembles may provide the best performance when trying to predict win probability. For Counter-Strike: Global Offensive (CSGO), Makarov~et~al.~\cite{makarov2017predicting} presented an ensemble approach using TrueSkill, decision trees and logistic regression to predict round winners. We see another CSGO player valuation framework in Bednarek~et~al.~\cite{bednarek2017player} which utilized the spatial information to cluster death locations to create player ratings. Specifically, they argue that encounters vary in importance by location and cluster deaths using k-means to create death heatmaps. Although the above esports works made good progress in valuing CSGO players, they lack a reproducible, context-aware player valuation framework.

%%%%% Data Processing %%%%%
\section{A Data Model for CSGO}
\label{sec:csgo-data}

\begin{figure*}
    \centering
    \includegraphics[width=0.9\linewidth]{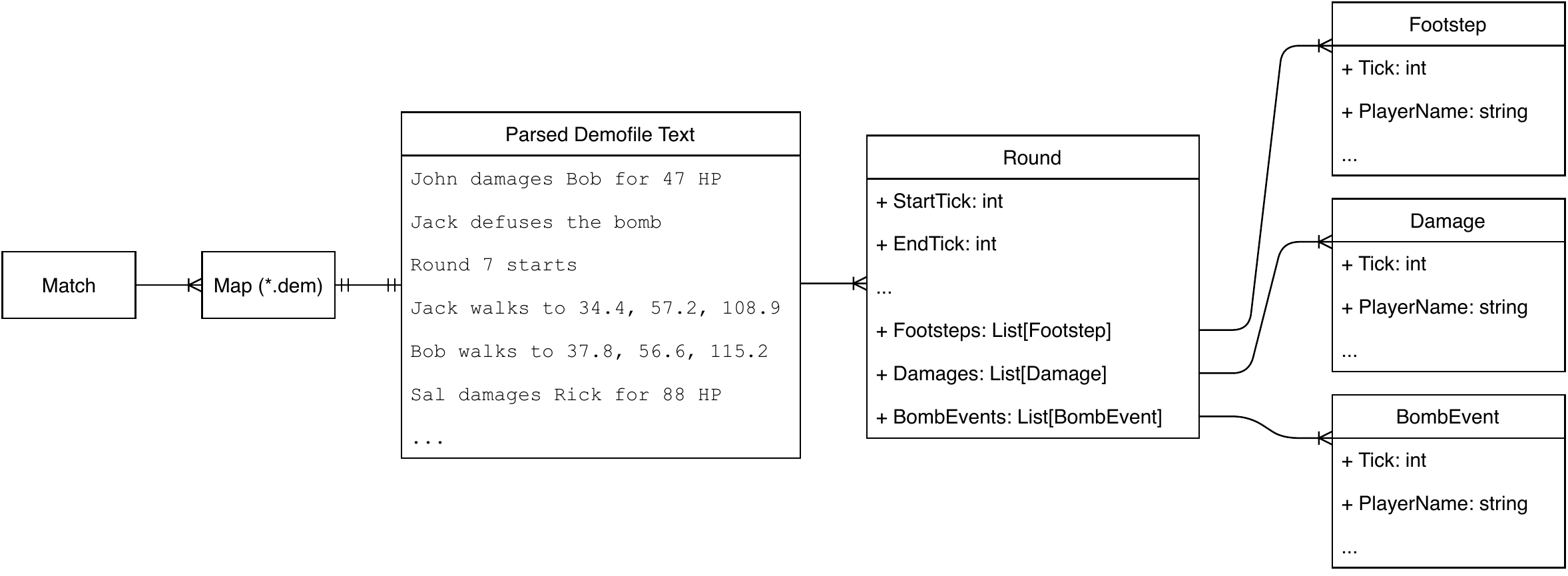}
    \caption{Unstructured demofile data is structured under an extensible model.}
    \label{fig:csgo-data-model}
\end{figure*}

\subsection{CSGO Game Description}
\label{sec:csgo-data-overview}
CSGO is a first-person shooter video game where two teams of 5 players compete to achieve a variety of objectives. These two teams play as both the Terrorist (T) and Counter-Terrorist (CT) sides over the course of a professional match. A professional match consists of a collection of performances on one or more \textit{maps}, which are distinct virtual worlds. Typically, competitions are structured as best of three or five maps. In professional competitions, maps are standardized and selected from a known pool of seven maps. On each map, teams are initially assigned T or CT and then play for 15 rounds as their assigned side. Teams switch sides after the 15th round. Whichever team wins 16 rounds out of 30 wins the map.

The T and CT sides can win a round through a variety of methods. Both teams can win a round if they eliminate the opposing side. Players start with 100 health points (HP) and are eliminated from the round when they reach 0 HP. Specifically, players lose HP when other players damage them via guns or grenades. Players buy equipment, such as guns, grenades and armor, at the beginning of a round, using money earned from doing well in previous rounds. Beyond eliminating the opposition, the T side can win a round by planting and exploding a bomb at one of two bombsites on a map, denoted A or B. One T player is randomly assigned the bomb at the start of each round, and this bomb can only be planted at one of the two bombsites. Once planted, the bomb explodes in 35 seconds, unless defused. The CT side can win the round by defusing a planted bomb.

\subsection{CSGO Data}

Each professional match typically generates recordings of the game called a demofile. Every map in a match will generate its own demofile. This demofile contains a serialization of the data transferred between the host (game server) and its clients (players). Data transferred to and from the server occurs at a predefined \textit{tick rate}, which defines when client inputs are resolved with the server. For professional games, the server tick rate is usually 128 ticks per second, meaning each tick represents around 7.8 milliseconds. Client inputs represent player actions, such as movement, attacks or grenade throws. Non-client events also register in the demofile, such as round starts and ends. For a detailed overview of CSGO demofiles, see Bednarek~et~al.~\cite{bednarek2017data}.

\subsection{CSGO Data Model}
\label{sec:csgo-data-model}

% One of the difficulties in working directly with CSGO demofiles is that since this is essentially data transferred between the clients and the game server, the data is simply stored as a text of sequential set of events, often with no contextual information, such as round or map location. 
Since CSGO demofiles are essentially data transferred between the clients and the game server, the data is simply stored as a text of sequential set of events 
% \harish{remove "often"?} 
with no contextual information, such as round or map location. 
Thus, due to the highly unstructured nature of these low level CSGO data streams, akin to log files, performing any complex analytic tasks becomes impossible without modeling this data formally into a useful data structure.
%
% One of the difficulties in parsing CSGO demofiles is that parser output is read as sequential strings of text, often with no contextual information, such as round or map location. 
% Thus, low level CSGO data streams are highly unstructured, and making joins on relevant events impossible without a predefined hierarchy. 
We therefore propose a hierarchical data model for CSGO data as illustrated in Figure~\ref{fig:csgo-data-model}.
% In Figure \ref{fig:csgo-data-model}, we present a data model for CSGO demofile data. First, a map's demofile is parsed into text. 
Given a specific map's demofile, we split this data into multiple \texttt{Round} objects, each of which contains relevant round information, such as start and end ticks, score, round results and events that happened during the round. All events that occurred during a round are stored in lists of objects corresponding to the type of event. For example, each \texttt{Round} object contains a list of \texttt{Footstep} objects that correspond to player movement events. Although we only detail an example using footstep, damage and bomb events, our data model is easily extensible to include other events such as grenade throws or shooting events.
% \harish{Peter: Please read the above paragraph to make sure I haven't messed anything. I've rephrased the text try and emphasize more the data modeling, and less the parsing.}

The events in \texttt{Round} objects also occur within a logical ordering based on their timestamp. These events progress the round through a series of \textit{game states}. We define a game state $G_{i,t}$ as an object that holds all of the current game information in round $i$ at tick $t$. A game state $G_{i,t}$ could contain attributes derived from round data, such as the map and score differential, temporal data such as the tick, spatial data such as player locations and other information such as the players remaining on each side or whether or not the bomb has been planted. When a player performs an action, such as damaging opponents or planting the bomb, the game advances from $G_{i,t}$ to $G_{i, t + 1}$. For purpose of this work, we define a game state with the attributes for each team outlined below:

\vspace{1mm}
\begin{description}
\item[Map:] the map where the game state occurred
\item[Ticks Since Start:] how many ticks since the round start
\item[Equipment Value:] the total round start equipment value
\item[Players Remaining:] the total numbers of players remaining
\item[HP Remaining:] the total health points remaining
\item[Bomb Planted:] a flag indicating if the bomb is planted
\item[Bomb Plant Site:] if the bomb is planted, at which site
\item[Team Bombsite Distance:] Minimum player distance to both bombsites for each side
\end{description}

While various CSGO demofile parsers exist, they simply read the demofiles as a sequence of text, thus making it difficult to integrate them with modern data science pipelines. Moreover, some existing parsers are also operating system specific, 
%lack relational data support 
or are written in programming languages without broad support in the data science community. 
On the other hand, the \texttt{Round} object, along with its corresponding events and game states, can each easily be stored as dictionaries or data frames, which make our data model congruent with contemporary data science workflows. Additionally, the data model is congruent to services that deliver JSON data, like many APIs. 
Our CSGO data model is implemented as Python library, which we will make available as an open source library so that it can be used by the growing esports analytics community.
% HD: removing the name \texttt{csgo} so people cannot explicitly search
%Our CSGO data model is implemented in the \texttt{csgo} Python library, which is available as an open source library, and can be used by the growing esports analytics community. 

\subsection{CSGO Spatial Data}
\label{sec:csgo-spatial-data}

Spatiotemporal data in contemporary sports is becoming valued as tracking systems become more reliable  \cite{gudmundsson2017spatio}. Despite this trend, player tracking data is still virtually unused in esports. Spatial data in sports is often used to create distance based metrics. For example, Yurko~et~al.~\cite{yurko2019going} uses many distance based features, such as distance to closest defender, to estimate the expected yards gained from an NFL ball carrier's current position. Similarly, Decroos~et~al.~\cite{decroos2019actions} uses distance based features, such as distance to the goal, for soccer goal prediction. For most sports, distance based metrics are based on euclidean distance, since sports like American football, soccer, basketball and baseball can be summarized on 2D surfaces with no obstructions. 

\begin{figure}
    \centering
    \includegraphics[width=\linewidth]{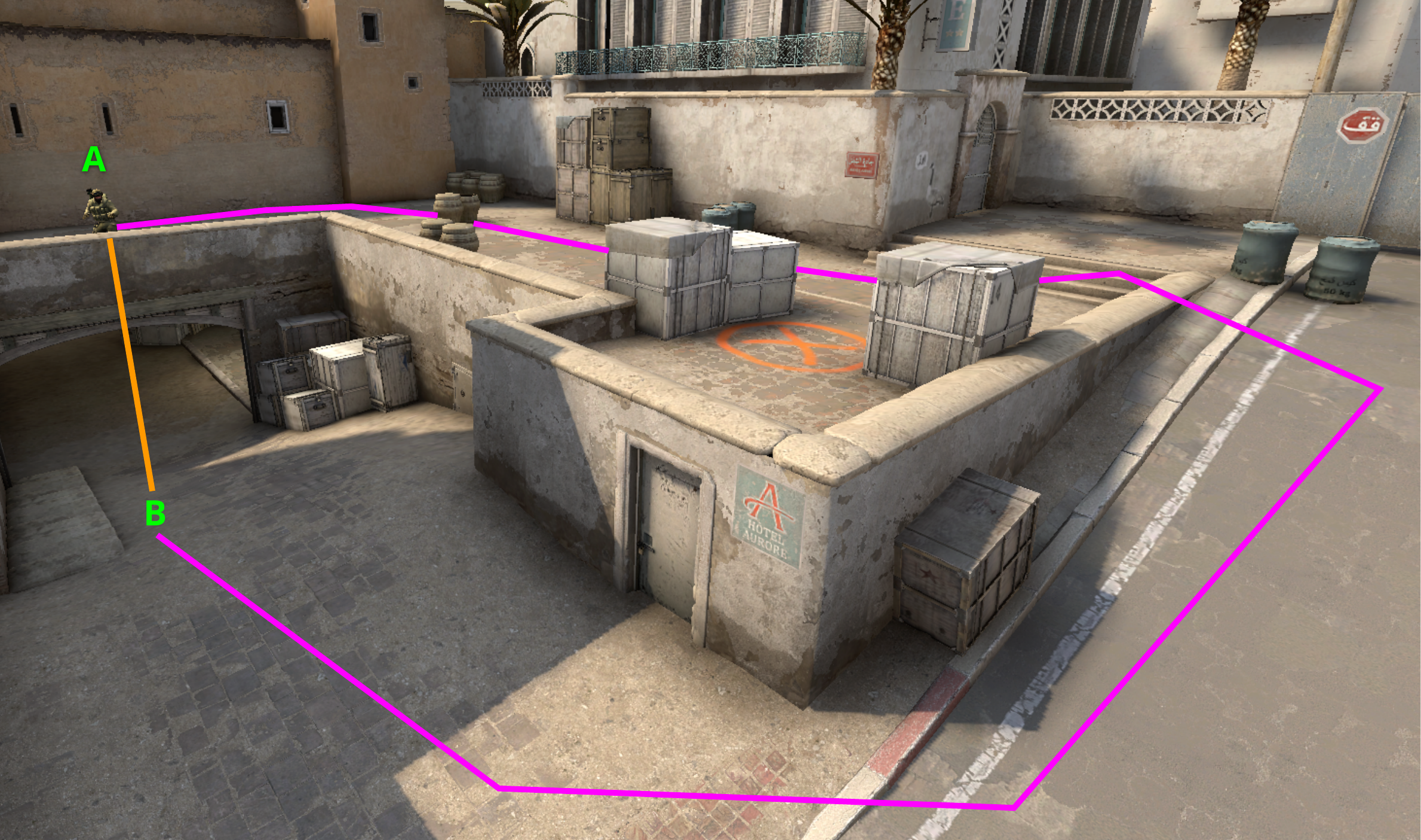}
    \caption{One can simply jump down from point A to point B via the orange path. However, to get from point B to point A, one must take the purple path, which illustrates the non-symmetric nature of distance in CSGO.}
    \label{fig:symmetry_example}
\end{figure}

Despite euclidean distance's ubiquity, it may not be useful in the spaces considered in esports. Firstly, esports maps often contain obstructions, like walls or buildings, where players are prohibited from moving. Secondly, because the world is 3D, distance symmetry doesn't always hold. For example, one might be able to jump down to a position, but not back up to where they once were. We illustrate the non-symmetric nature of distance in CSGO in Figure~\ref{fig:symmetry_example}. To address these issues, we propose a graph-based distance measure which uses a graph that discretizes a CSGO map. 

\begin{figure}
    \centering
    \includegraphics[scale=0.45]{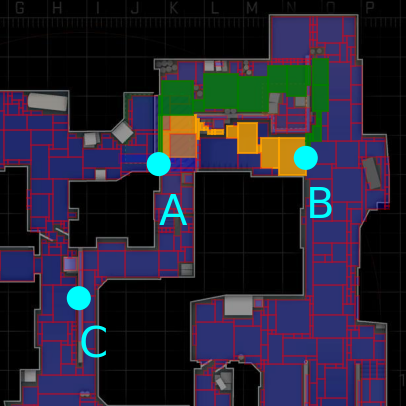}
    \caption{Although B and C are equidistant in euclidean space to A, A-to-C is closer in graph distance (13 nodes) than A-to-B (15 nodes in orange). Additionally, the green path (19 nodes) from B-to-A shows the non-symmetry of graph distance, since the map is 3D. The black space between C and B represent off limits/walled off areas. Each traversable surface is indicated as a blue shaded region with a red border.}
    \label{fig:distance_example}
\end{figure}

Although there are infinitely many ways to discretize the 3D map space in CSGO, we take an approach drawing from AI-controlled bot movement. A bot is an AI-controlled player which plays against real players, and CSGO provides the functionality for players to play against bots. Bots use \textit{navigation meshes}, which provide information on all traversable surfaces of the map, to move around the map~\cite{navigation}. Part of the information included for each surface includes all neighboring surfaces. From the navigation mesh, we can create a directed graph that represents the map. Each node represents a surface in the navigation mesh, while each directed vertex from node $a$ to node $b$ denotes a path available from $a$ to $b$, given $a$ and $b$ are adjacent. Each player at any given time in a round can be assigned a node in the graph depending on which surface they occupy. Using the $A^*$ algorithm, we can then find the shortest path from one node to another, which we call \textit{graph distance}. We show an example of graph distance in Figure~\ref{fig:distance_example}. We see that although both C and A are equidistant to B using euclidean distance, the graph distances reveal that A (via the green tiles) is about twice as close to B than is C (via the orange tiles). Furthermore, we see that B is closer to C (via the purple path) than C is to B. This is because of a height difference in the map that allows B to jump down to C, whereas C must take the long way around. 

%\begin{figure}
%    \centering
%    \includegraphics[scale=0.70]{figures/csgo_distance.pdf}
%    \caption{Although C and A are equidistant in euclidean space to B, A is closer in %graph distance (8 nodes in green) than C (19 nodes in orange). Additionally, the purple path (10 nodes) from B to C shows the non-symmetry of graph distance, since the map is 3D. The black space between C and B represent off limits/walled off areas.}
%    \label{fig:distance_example}
%\end{figure}

\begin{figure*}
    \centering
    \includegraphics[width=\textwidth]{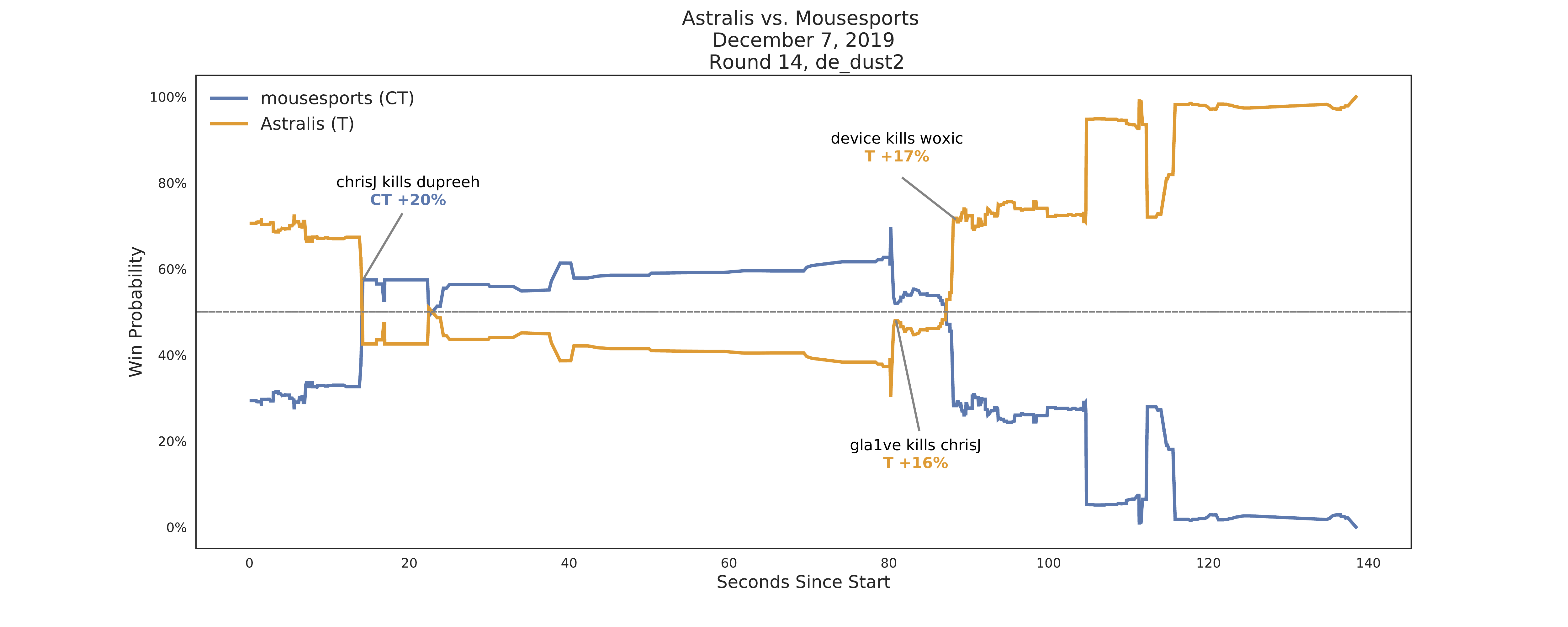}
    \caption{A team's win probability changes as players conduct actions such as movement, damages and bomb plants/defuses. Players are credited for their actions' increase or decrease to their team's win probability.}
    \label{fig:win-prob-example}
\end{figure*}
\section{Valuing Players}
\label{sec:valuation-framework}

% makarov did before bomb plants
\subsection{Existing Frameworks}
\label{sec:existing-frameworks}
Although CSGO analytics literature is sparse, there exist a few basic player valuation frameworks. For example, the simplest way to value a player is by using some function of their total kills and deaths. Intuitively, good players will have more kills than deaths. While we can simply tabulate a player's raw number of kills or deaths, we can also find their \textit{kill-death ratio}, defined as \begin{equation}
    KDR = \frac{Kills}{Deaths}
\end{equation}
One of the immediate drawbacks of valuing players using KDR is that players are only rewarded for attaining a kill and not for damaging other players or getting assists. 

Two metrics that attempt to address KDR's shortcomings include average damage per round (ADR) and KAST\% ~\cite{milanovic_2017}. ADR is simply a player's total damage output over rounds played. One of the downsides of ADR is that damage dealt may depend heavily on contextual factors such as the number of players remaining on each team or equipment value. KAST\% measures the proportion of rounds that a player achieves a beneficial event, defined as a \textbf{k}ill, \textbf{a}ssist, \textbf{s}urvival or \textbf{t}rade. Some drawbacks to KAST\% include all four events being valued the same, along with inconsistent definitions across the CSGO community for a ``trade'', which is loosely defined when a player kills another player but is shortly killed after. Furthermore, a player can attain a perfect KAST\% score by purposely disengaging with enemies the entire match. ADR and KAST\% are calculated as 
\begin{eqnarray}
    % \begin{split}
        ADR &= \frac{\text{Total Damage}}{\text{Rounds}} \\
        KAST\% &= \frac{ \text{Kills} + \text{Assists} + \text{Survivals} + \text{Trades} }{\text{Rounds}}
    % \end{split}
\end{eqnarray}

To create a more comprehensive player valuation metric, the popular CSGO website HLTV developed HLTV Rating 1.0 and 2.0. HLTV Rating 1.0 is defined as 
\begin{equation}
    Rating_{1.0} = \frac{Rating_{K} + 0.7 \times Rating_{S} + Rating_{MK}}{2.7}
\end{equation}
where $Rating_K$, $Rating_S$ and $Rating_{MK}$ are a player's kill rating, survival rating and multiple kills rating, respectively. These ratings are functions of a player's kills and deaths~\cite{milanovic_2010}. To account for kills being worth more than survivals, Rating 1.0 weighs $Rating_S$ less. Although HLTV's Rating 1.0 methodology was public, the Rating 2.0 system methodology is not. HLTV's Rating 2.0 includes more ratings, such as a KAST and Damage rating, along with different rating calculations for both T and CT performances~\cite{milanovic_2017}. While HLTV Rating 2.0 takes a step in the right direction by calculating separate T and CT ratings, its exact mechanisms are still unknown making it impossible to reproduce. Furthermore, beyond controlling for side, additional features can influence the difficulty of game situations. To address the shortcomings of the aforementioned player valuation systems, we introduce a context-aware player valuation framework based on game state transitions.

\subsection{Valuation Framework}
Over a course of a round, player actions transition the game through a sequence of game states $G_{i,1}, ..., G_{i,t}$, as described in Section~\ref{sec:csgo-data-model}. Each game state is also associated with a round outcome, $Y_i$, which is coded as 1 if the CT sides wins round $i$ and 0 otherwise. We are interested in estimating $P(Y_i = 1 \mid G_{i,t})$. Let $\widehat{Y}_{i,t}$ be the estimated win probability given a game state $G_{i,t}$. Thus, $\widehat{Y}_{i,t}$ changes based on the game state, and can change drastically depending on the game scenario. We can see an example of how win probability changes according to player actions in Figure~\ref{fig:win-prob-example}. 

Now, assume a player committed an action in round $i$ at time $t$, denoted as $a_{i,t}$. Although this action could be any event, such as a footstep or bomb plant, we only define damage events as actions for our framework. Once action $a_{i,t}$ has occurred, the game state transitions from $G_{i,t}$ to $G_{i,t+1}$, and our estimate of $Y_i$ changes as well. We can value the player's action as the difference in win probability between the two states, formally defined as

\begin{equation}
    V(a_{i,t}) = \widehat{Y}_{i, t+1} - \widehat{Y}_{i,t} \label{eq:value}
\end{equation}

It is clear that for any action, $V(a_{i,t}) \in [-1, 1]$. Since $Y_i$ is defined as a CT win, any action beneficial to the T side will return a negative value. Because we want to credit players on both T and CT positively for beneficial actions to their teams, we normalize $V(a_{i,t})$ to the team of the player committing action. The player on the receiving side of the action (e.g. the player receiving damage) is credited with the negative of $V(a_{i,t})$. To determine a player's total contribution, we can tabulate their \textit{Win Probability Added} (WPA), defined as the sum of the player's total action values over all games. To standardize the metric, we report WPA as WPA per round, since games have a variable number of rounds. 

\subsection{Estimating Win Probability}
\label{sec:win-prob}
Previous work on estimating win probability formulated the problem as a classification task and utilized methods such as logistic regression~\cite{yurko2019nflwar, macdonald2012expected, yang2016real}, tree based classifiers~\cite{decroos2019actions, hodge2019win} or neural networks~\cite{liu2018deep, yurko2019going}. We use a similar set up, since estimating $Y_{i}$ conditioned on $G_{i,t}$ presents a classic binary classification problem. We structure our observations as a collection of game states, where our features are the game state attributes described in Section~\ref{sec:csgo-data-model}.

Our data consists 4,682 local area network (LAN) matches that contain public demofiles. We use LAN matches because they are typical professional matches played in a tournament setting. We downloaded the demofiles used in our study, along with each matches' ADR, KAST and HLTV Rating 2.0 from HLTV, one of the most popular CSGO websites that contains news, match statistics and demofiles. We construct a training set of 55 million game states from matches from October 23rd, 2016 to May 31st, 2019 and a test set of 18 million game states from matches occurring from June 1st, 2019 to December 22nd, 2019. We observe a game state whenever a footstep, damage or bomb event occurred. However, to calculate WPA, we only consider damage events.

Since WPA is highly dependent on obtaining good estimates for $Y_i$, we focus on a selection of models that can produce well calibrated probabilities. Specifically, we consider logistic regression, CatBoost~\cite{prokhorenkova2018catboost} and XGBoost~\cite{chen2016xgboost}. We focus on boosted ensembles for their demonstrated tendency to produce well calibrated probabilities~\cite{niculescu2005obtaining, niculescu2005predicting}. As we are interested in estimating probabilities, we define performance using both the log loss and the Brier loss, which are standard in probabilistic prediction problems, as well as AUC~\cite{vovk2015fundamental}. Additionally, we consider a baseline model which predicts a game state's map average CT win rate. We outline our model tuning procedures and computing environment below.

\subsection{Hyperparameter Tuning}
\label{appendix:hyperparameters}
We trained all of our models on a server running Ubuntu 16.04 with 2x Intel Xeon E5-2695 2.4 GHz, 256GB of RAM and 3 NVIDIA Titan GPUs. We utilized the \textit{scikit-learn} implementation of logistic regression with the SAGA solver and no regularization, along with the \textit{catboost} and \textit{xgboost} packages for our CatBoost and XGBoost implementations. To tune our parameters, we used a train/validation split where $\frac{2}{3}$ of the data was used for training and $\frac{1}{3}$ was used to validate our models. Unless stated, we left the parameters set to their defaults. We used a log loss scoring function.

\subsubsection{XGBoost Tuning}
We performed a grid search across the parameters shown in Table~\ref{tab:appendix-params}. The optimal parameter set is in bold. We set the number of estimators to $100$. Our final model reported in the paper is the one chosen by the search. We used the \texttt{hist} tree method~\cite{xgboost_params}. 

\subsubsection{CatBoost Tuning}
We performed a grid search across the parameters shown in Table~\ref{tab:appendix-params}. The optimal parameter set is in bold. We set the number of iterations to $100$. Our final model reported in the paper is the one chosen by the search. We trained our model using all available GPUs.

\begin{table}[h]
\centering
\caption{XGBoost and CatBoost Parameter Spaces}
\begin{tabular}{@{}ccc@{}}
\toprule
\textbf{Model} & \textbf{Parameter} & \textbf{Values} \\ \midrule
XGBoost &  &  \\
 & \texttt{max\_depth} & 6, \textbf{8}, 10, 12 \\
 & \texttt{min\_child\_weight} & \textbf{1}, 3, 5, 7 \\ \midrule
CatBoost &  &  \\
 & \texttt{depth} & 6, 8, \textbf{10}, 12 \\
 & \texttt{l2\_leaf\_reg} & 1, \textbf{3}, 5, 7 \\ \bottomrule
\end{tabular}

\label{tab:appendix-params}
\end{table}

\begin{figure}[t]
    \centering
    \includegraphics[width=\linewidth]{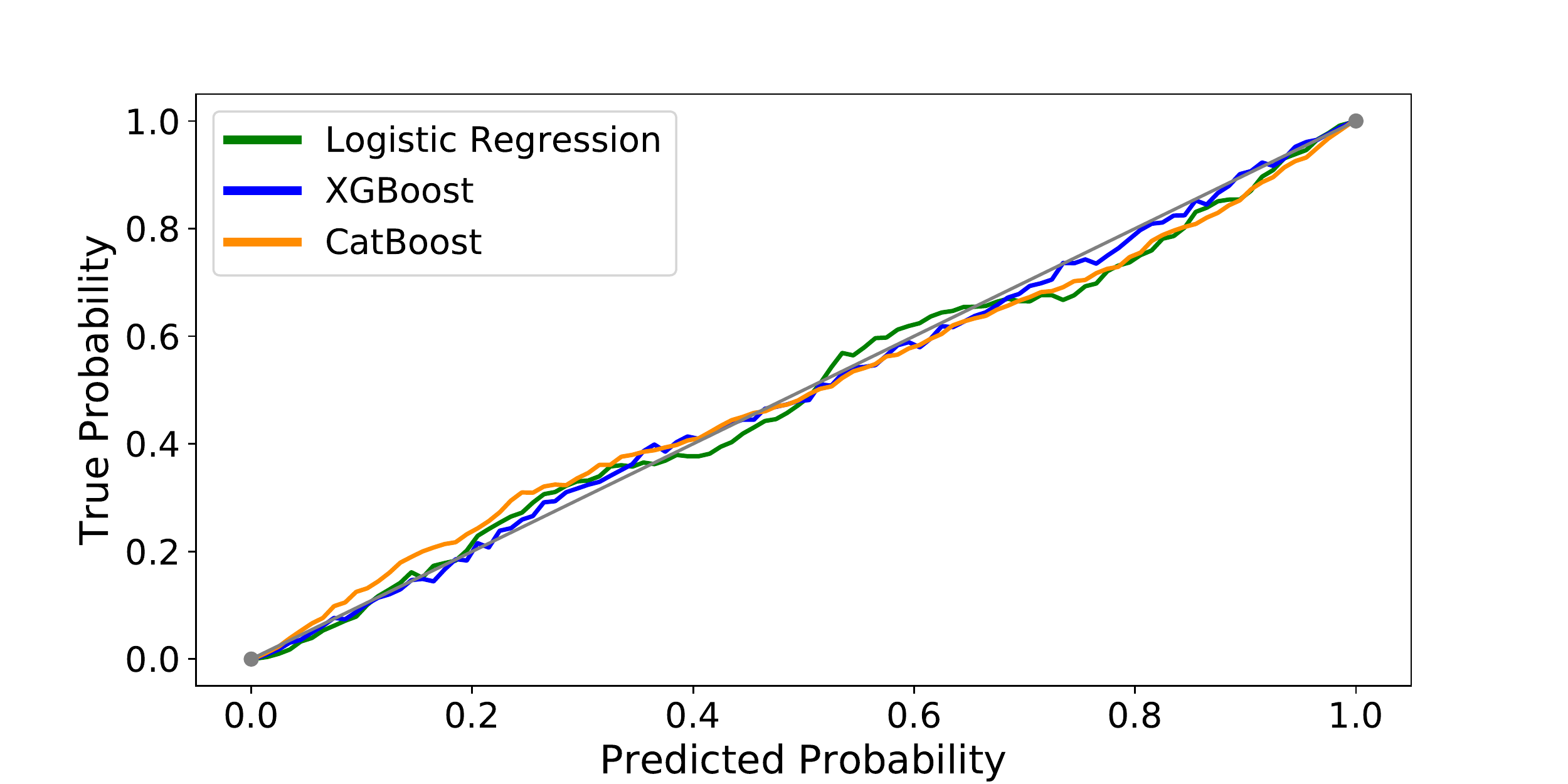}
    \caption{Calibration curves on test data for each model show that XGBoost produced the most well calibrated win probabilities.}
    \label{fig:model-calibration}
\end{figure}

\section{Model Results}
\label{sec:results}

\subsection{Model Performance}
\label{sec:model-performance}
As Decroos~et~al.~\cite{decroos2019actions} note, assessing the performance of a player valuation system is difficult as there exists no ground truth for player valuation. Therefore, we first assess our system through the performance of our win probability model and the insights the model provides. Then, we provide a comprehensive comparison of WPA against other metrics along with different use cases of the metric. We begin by assessing the performance of our win probability model. We present a performance comparison of logistic regression, CatBoost and XGBoost for predicting the round winner in Table~\ref{tab:algorithm-performance}. We see that all models perform significantly better than the map average benchmark, and that XGBoost performs the best across all metrics.

\begin{table}[h]
\centering
\caption{Win probability model performance on test data. XGBoost outperformed all other candidate models across all metrics.}
\label{tab:algorithm-performance}
\begin{tabular}{@{}cccc@{}}
\toprule
\textbf{Method} & \textbf{Log Loss} & \textbf{Brier Score} & \textbf{AUC} \\ \midrule
Logistic Regression & 0.5539 & 0.1912 & 0.7743 \\ 
CatBoost & 0.5443 & 0.1875 & 0.7851  \\ 
XGBoost & \textbf{0.5353} & \textbf{0.1842} & \textbf{0.7913}  \\ \midrule
Map Average & 0.6917 & 0.2493 & 0.5303 \\ \bottomrule
\end{tabular}
\end{table}

\subsection{Model Calibration}
\label{results:model-calibration}
Beyond comparing model performance in log loss, Brier score or AUC, it is also important to know if our models are producing well calibrated probabilities, since these probabilities are crucial for calculating WPA. We present a calibration plot of our three models in Figure~\ref{fig:model-calibration}. We generate this calibration plot by creating 100 equal width bins of predictions, where we plot each bin's mean predicted probability on the x-axis and the bin's mean true probability on the y-axis. While we observe that all of our models were well calibrated, XGBoost seemed to produce the most well calibrated probabilities, as it most closely follows the perfect calibration line in gray. 

\begin{figure}[t]
    \centering
    \includegraphics[width=\linewidth]{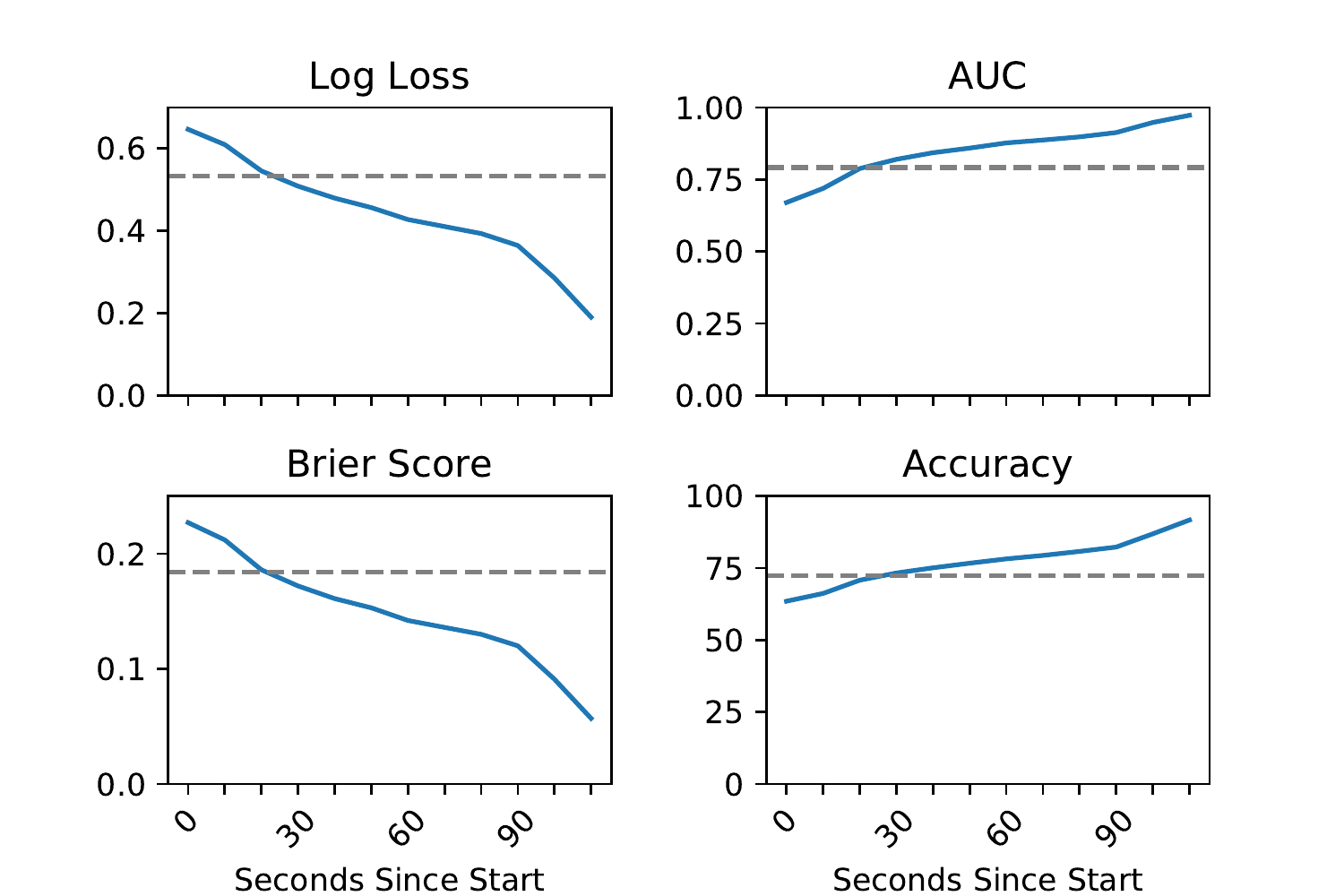}
    \caption{Win probability model performance by round time indicates that our win probability model becomes more powerful later in rounds.}
    \label{fig:model-performance}
\end{figure}

Another way to assess our win probability model is to investigate its performance at different timestamps in a round. We would expect our model to predict the winner of a round much better closer to the end of a round than at the beginning, since the time horizon to the round end is smaller. In Figure~\ref{fig:model-performance}, we see that, as expected, log loss and Brier score decrease, while AUC and accuracy increase as the rounds progress. This finding gives our win probability model utility in applications such as live betting or media coverage of an event, aside from its use in valuing CSGO players. One criticism of the model training methodology is that many of the training examples are highly correlated, such as adjacent game states. However, given that the test data is completely independent from the training data, along with both intuitive and strong model performance, we see that our model still generalizes well. 

\subsection{Feature Importance}
\begin{figure}[t]
    \centering
    \includegraphics[width=\linewidth]{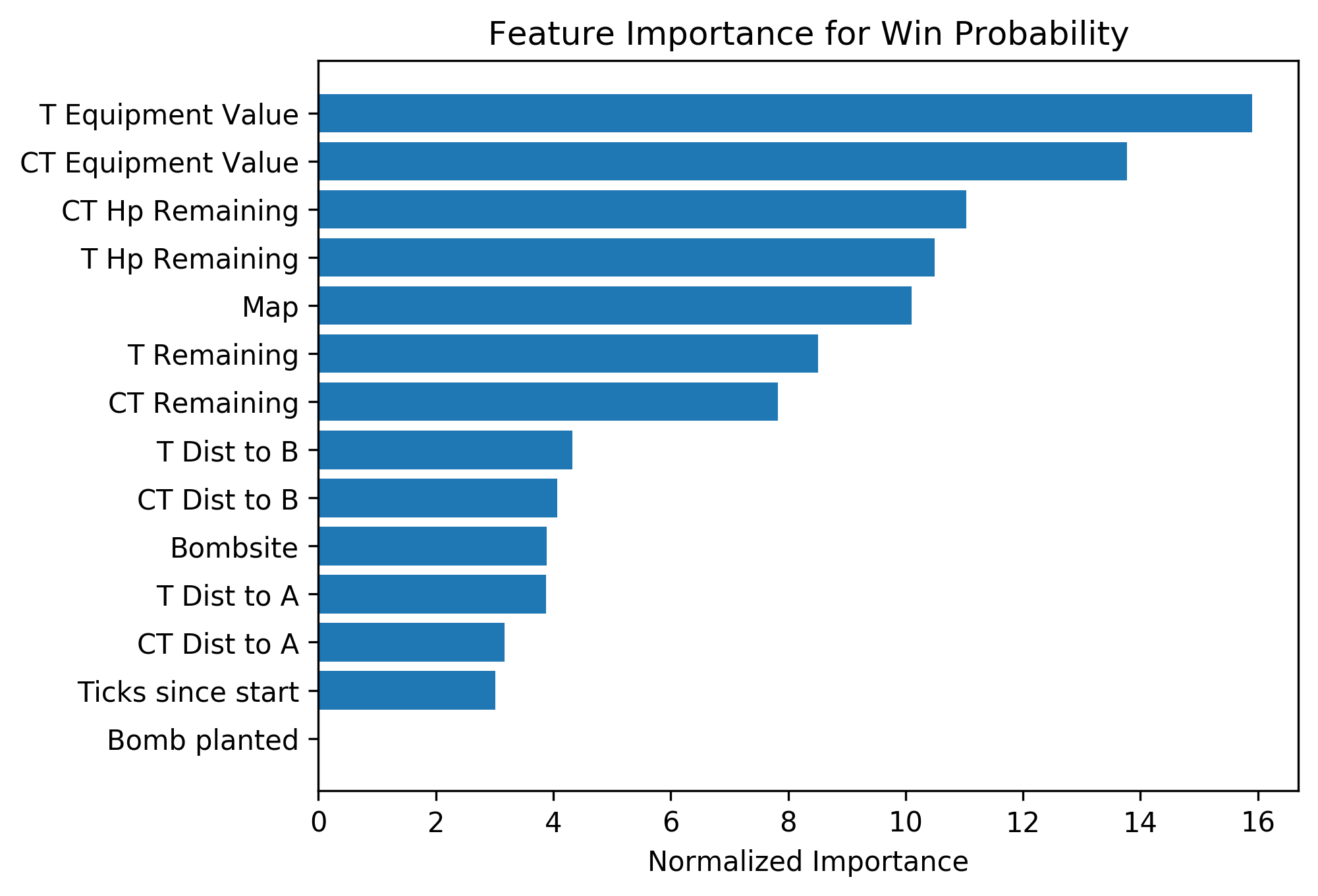}
    \caption{Intuitively, team equipment and HP remaining had the highest feature importance. Additionally, the map the round took place on was also quite important, highlighting its interaction effects with the other features.}
    \label{fig:feature-importance}
\end{figure}

One of the benefits of the XGBoost algorithm is that it provides feature importances through analyzing the total gain provided by each feature. Understanding what features are driving the model can help teams consider relevant aspects of their gameplay. In Figure~\ref{fig:feature-importance}, we show the normalized importances for each feature in our win probability model. The scores are normalized so the sum of all importances is 100. 

From Figure~\ref{fig:feature-importance}, it is apparent that a team's equipment value is the lead determinant of a team's chances of winning. This is intuitive, since teams gain a significant advantage in the game when they have better equipment. We then see that HP remaining for both sides is another strong factor for predicting what side wins a round. What is interesting is that these factors rank higher than the players remaining, which may suggest that while having a numerical advantage is important, the team health may be more important for winning a round. Ultimately, these features can change win probability drastically, as seen Figure~\ref{fig:win-prob-example}. 
In this particular example, note that the kill events drive the majority of win probability changes.
% we can visually inspect the consequences of these features, where we notice that kill events drive the majority of win probability changes.

Another interesting consequence from the feature importance plot is the nature of the \textit{Map} feature. While solely using the Map feature leads to significantly worse prediction performance, as indicated in Table~\ref{tab:algorithm-performance} under \textit{Map Average}, the feature is still ranked fifth in feature importance. This relationship suggests that the Map feature provided important context in the presence of other features. This finding also suggests that there may exist optimal map strategies under certain team equipment, HP and players left constraints.

\section{Discussion}
\label{sec:discussion}

\subsection{Comparison with Existing Frameworks}
\label{sec:comparison-frameworks}

While WPA certainly accounts for deeper context than existing metrics through an underlying reproducible, interpretable and performant model, it is also important to assess the usefulness of WPA as a sports metric. Franks~et~al.~\cite{franks2016meta} suggest that a useful sports metric exhibits the following three traits: 

\begin{description}
\item[Stability:] The metric's ability to measure the same quantity over time
\item[Discrimination:] The metric's ability to differentiate players
\item[Independence:] The metric's ability to provide new information
\end{description}

In the following analysis, we focus on stability and independence. If a new metric is too closely correlated with existing measures, it could be measuring the same underlying quantity as the existing measures, rendering the new metric useless. Additionally, if a new metric doesn't give consistent measurements of player, it may not be adopted by key stakeholders, such as teams, media or gamblers.

\begin{table*}[t]
\centering
\caption{Top 10 for all rounds and pistol rounds only. Discrepancies between WPA and HLTV rankings further denote the uniqueness of WPA compared to traditional metrics.}
\label{tab:top-ten}
\begin{tabular}{ccccllccc}
\multicolumn{3}{c}{\textit{All Rounds}} &  &  &  & \multicolumn{3}{c}{\textit{Pistol Rounds}} \\ \cline{1-3} \cline{7-9} 
\multicolumn{1}{|c}{\textbf{Player}} & \textbf{WPA} & \multicolumn{1}{c|}{\textbf{HLTV Rank}} &  &  & \multicolumn{1}{l|}{} & \textbf{Player} & \textbf{WPA} & \multicolumn{1}{c|}{\textbf{HLTV Rank}} \\ \cline{1-3} \cline{7-9} 
\multicolumn{1}{|c}{ZywOo} & 0.044 & \multicolumn{1}{c|}{1} &  &  & \multicolumn{1}{l|}{} & cadiaN & 0.080 & \multicolumn{1}{c|}{9} \\
\multicolumn{1}{|c}{KSCERATO} & 0.033 & \multicolumn{1}{c|}{27} &  &  & \multicolumn{1}{l|}{} & ZywOo & 0.068 & \multicolumn{1}{c|}{11} \\
\multicolumn{1}{|c}{s1mple} & 0.028 & \multicolumn{1}{c|}{3} &  &  & \multicolumn{1}{l|}{} & EliGE & 0.067 & \multicolumn{1}{c|}{40} \\
\multicolumn{1}{|c}{acoR} & 0.027 & \multicolumn{1}{c|}{36} &  &  & \multicolumn{1}{l|}{} & electronic & 0.057 & \multicolumn{1}{c|}{32} \\
\multicolumn{1}{|c}{woxic} & 0.025 & \multicolumn{1}{c|}{31} &  &  & \multicolumn{1}{l|}{} & huNter- & 0.051 & \multicolumn{1}{c|}{7} \\
\multicolumn{1}{|c}{ropz} & 0.025 & \multicolumn{1}{c|}{21} &  &  & \multicolumn{1}{l|}{} & shox & 0.051 & \multicolumn{1}{c|}{60} \\
\multicolumn{1}{|c}{xsepower} & 0.022 & \multicolumn{1}{c|}{9} &  &  & \multicolumn{1}{l|}{} & KSCERATO & 0.048 & \multicolumn{1}{c|}{26} \\
\multicolumn{1}{|c}{device} & 0.022 & \multicolumn{1}{c|}{7} &  &  & \multicolumn{1}{l|}{} & Brehze & 0.045 & \multicolumn{1}{c|}{10} \\
\multicolumn{1}{|c}{EliGE} & 0.021 & \multicolumn{1}{c|}{5} &  &  & \multicolumn{1}{l|}{} & dexter & 0.042 & \multicolumn{1}{c|}{6} \\
\multicolumn{1}{|c}{Jame} & 0.021 & \multicolumn{1}{c|}{14} &  &  & \multicolumn{1}{l|}{} & device & 0.042 & \multicolumn{1}{c|}{28} \\ \cline{1-3} \cline{7-9} 
\end{tabular}
\end{table*}

WPA should act as a reliable measure of player talent in that it should measure the same quantity over time. That is, if we take multiple measures of a player's talent, the sampled measures are consistent. To compare WPA's stability to existing metrics, we calculated the month-to-month correlation for the various existing measures and WPA. We use a month-to-month correlation as it is unlikely that a player's talent level changes significantly in such a short time frame. Furthermore, a month-to-month design is similar to the season-to-season approach in~\cite{franks2016meta}. Using players in our dataset who played at least 100 rounds each month from June 2019 to December 2019 $(n = 479)$, we show the month-to-month correlation across all metrics in Table~\ref{tab:correlation}. Using the Fisher r-to-z transformation to assess the significance of difference between WPA's month-to-month correlation, we see that it attains significantly greater stability than ADR ($p = 0.0029)$, KAST\% ($p = 0.0392$) and Rating 2.0 ($p = 0.0268$), where the parenthesis report one-sided difference-in-correlation test p-values~\cite{fisher1915frequency}. WPA's stability improvement over KDR's was not statistically significantly ($p = 0.3594$). We see that the evidence supports the notion that WPA is more stable than many of the current advanced CSGO player metrics today.

One of the chief downsides of popular CSGO player valuation metrics is that they correlate highly with kill-death ratio (KDR). Thus, many existing CSGO metrics violate the independence trait described above. We observe the correlation between KDR, existing valuation metrics and WPA in Table~\ref{tab:correlation}. Using the same Fisher r-to-z transformation methodology as above, we report the following one-sided p-values to show that WPA is not more independent with KDR than ADR ($p = 0.9656$), but more independent with KDR than KAST\% ($p = 0.0139$) and HLTV Rating 2.0 ($p = 0.0000$). The difference in correlation can be attributed to the consideration for the relevant context in which kills and damages occur. We believe the correlation between WPA and existing metrics can be further lowered by considering non-damage player actions, such as bomb plants, defuses and movement. Our results give us confidence that WPA is a useful sports metric since, unlike existing metrics, it provides a more consistent measure of a quantity that is less correlated with KDR. 

\begin{table}[h]
\centering
\caption{WPA attains the highest month to month (M-to-M) correlation, indicating its stability. WPA also is significantly less correlated with KDR than Rating 2.0, demonstrating its higher degree of independence as a metric.}
\label{tab:correlation}
\begin{tabular}{@{}ccc@{}}
\toprule
\textbf{Metric} & \textbf{M-to-M} & \textbf{KDR} \\ \midrule
KDR & 0.38 &  1.00 \\
ADR & 0.24 &  0.67  \\
KAST\% & 0.30 &  0.79  \\
Rating 2.0 & 0.29 &  0.93  \\ \midrule
\textit{WPA} & \textit{0.40}  & \textit{0.73} \\ \bottomrule
\end{tabular}
\end{table}

While we have shown that WPA is both a consistent and independent metric when compared to other CSGO player metrics, it is important to verify that it produces intuitive results. In Table~\ref{tab:top-ten}, we present the top 10 players using WPA per round. While there is some agreement between WPA and HLTV Ranking, players outside of the standard top 10 are included in WPA's top 10. Seeing well-accepted names give credence in WPA's ability to detect good players. However, the discrepancies also further highlight how WPA is independent of HLTV Rating 2.0, which is strongly correlated with KDR. 

One of the benefits of WPA is that we can calculate WPA for a variety of scenarios. We can calculate WPA for certain round contexts by filtering game states on certain attributes such as players remaining or win probability. For example, the pistol round is considered an important round, as it occurs on the first and 16th rounds. In this round, players on each team start with only pistols. Winning the pistol round allows the winning team to gain a significant money advantage in game. With this in mind, we calculate the WPA per round on pistol rounds for our data and present the results in Table~\ref{tab:top-ten}. It is clear that there is far more variation, supporting WPA's independence from traditional metrics like HLTV Rating and KDR.

\hidecomment{
\begin{table}[b]
\begin{tabular}{@{}cccc@{}}
\toprule
\textbf{Player} & \textbf{WPA} & \textbf{Rating 2.0} & \textbf{HLTV Ranking} \\ \midrule
ZywOo & 0.044 & 1.33 & 1 \\
KSCERATO & 0.033 & 1.17 & 27 \\
s1mple & 0.028 & 1.30 & 3 \\
acoR & 0.027 & 1.13 & 36 \\
woxic & 0.025 & 1.21 & 31 \\
ropz & 0.025 & 1.16 & 21 \\
xsepower & 0.022 & 1.19 & 9 \\
device & 0.022 & 1.26 & 7 \\
EliGE & 0.021 & 1.18 & 5 \\
Jame & 0.021 & 1.20 & 14 \\ \bottomrule
\end{tabular}
\caption{Top 10 WPA per Round}
\label{tab:top-players}
\end{table}

\begin{table}[t]
\begin{tabular}{@{}ccc@{}}
\toprule
\textbf{Player} & \textbf{WPA} & \textbf{HLTV Ranking} \\ \midrule
cadiaN & 0.080 & 9 \\
ZywOo & 0.068 & 11 \\
EliGE & 0.067 & 40 \\
electronic & 0.057 & 32 \\
huNter- & 0.051 & 7 \\
shox & 0.051 & 60 \\
KSCERATO & 0.048 & 26 \\
Brehze & 0.045 & 10 \\
dexter & 0.042 & 6 \\
device & 0.042 & 28 \\ \bottomrule
\end{tabular}
\caption{Top 10 WPA per Pistol Round}
\label{tab:pistol-ranking}
\end{table}
}

\subsection{Uncertainty Estimation}
\label{discussion:uncertainty}

\begin{figure}[t]
    \centering
    \includegraphics[width=\linewidth]{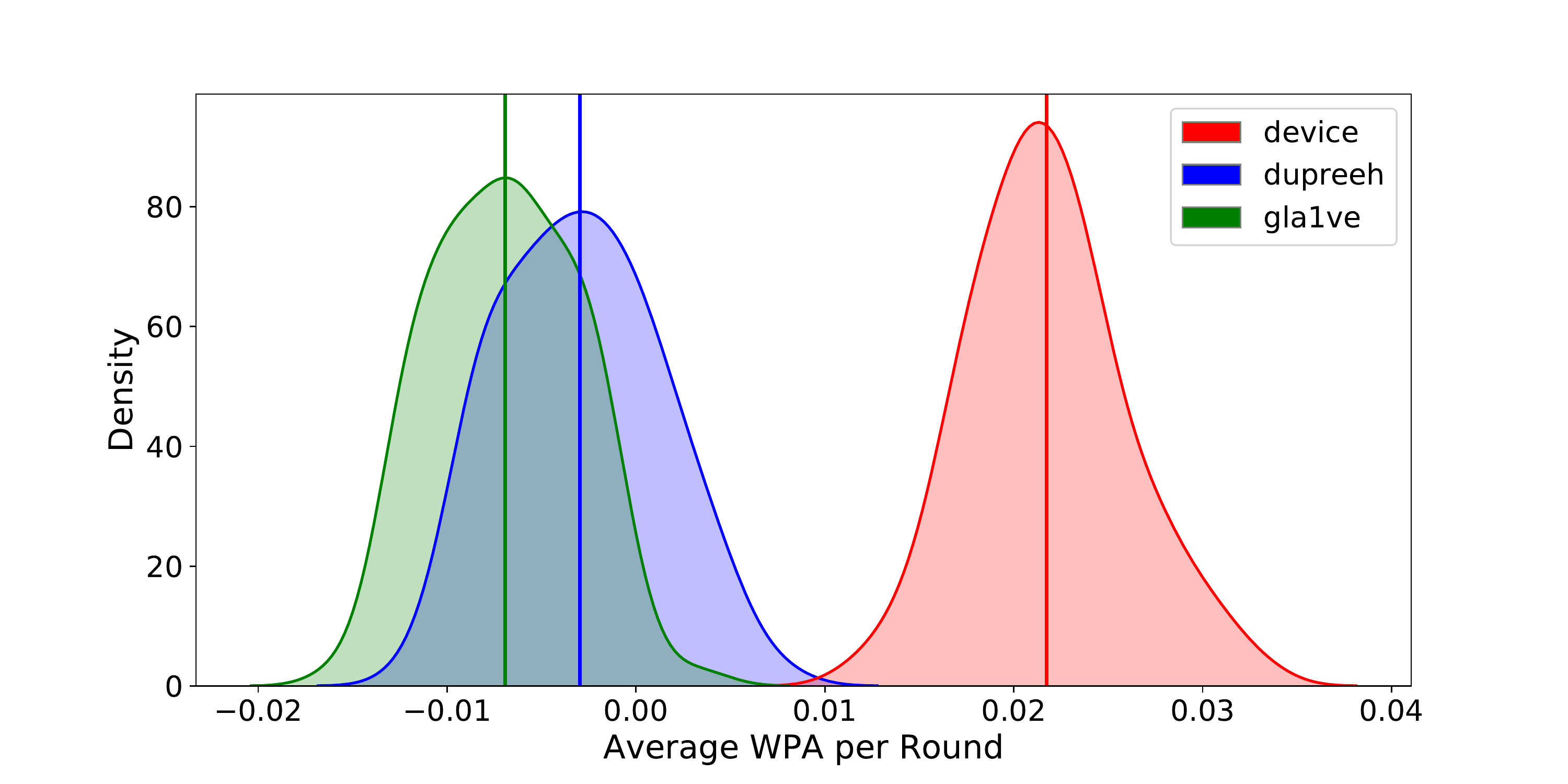}
    \caption{WPA per round estimates for \texttt{device} (red), \texttt{dupreeh} (blue) and \texttt{gla1ve} (green). Although \texttt{gla1ve} has the worst average WPA per round, he also has the lowest variance.}
    \label{fig:uncertainty}
\end{figure}

\begin{figure*}[t]
    \centering
    \includegraphics[width=\textwidth]{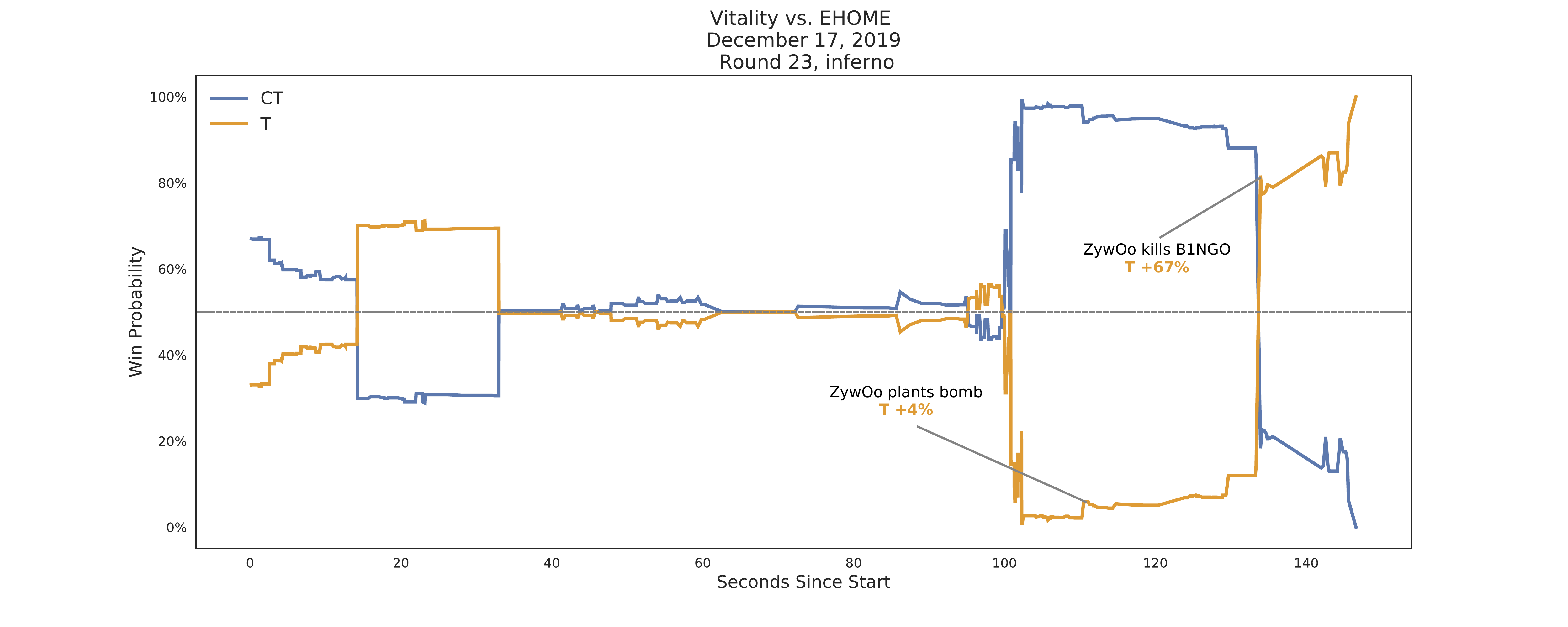}
    \vspace{-0.3in}
    \caption{A 67\% win probability increase due to a kill in a 1-vs-2 situation by \texttt{ZywOo} demonstrates a high impact play which is easily discovered through using win probability.}
    \label{fig:big-play}
    \vspace{-0.1in}
\end{figure*}

Player valuation frameworks typically only report a point estimate. However, understanding the variability of estimates is important for a variety of stakeholders. For example, media and fans may speculate on the upside of players, teams may be concerned with finding consistent players and bettors may be want to limit risk in their betting portfolio. However, even if we knew a player's true talent, there would still be variation in the player's outcomes. Similar to the work in~\cite{baumer2015openwar, yurko2019nflwar}, we use a resampling strategy to understand player outcome variability by generating distributions of each player's mean WPA per round. Specifically, we resample each round a player has taken part in, with replacement. Then, we calculate a player's WPA per round.

To illustrate our uncertainty estimation procedure, consider the following three players: \texttt{device}, \texttt{dupreeh} and \texttt{gla1ve}. These three are members of Astralis, a top-tier CSGO team. In Figure~\ref{fig:uncertainty}, we plot their resampled mean WPA per game from games occurring between June 1st, 2019 and December 22nd, 2019 using 100 bootstrapped samples. Specifically, we resampled events that occurred between June 1st, 2019 to December 22nd, 2019. It is clear from our distributional estimates that there is a substantial difference between each of the players. For example, the players each exhibit different variances. We see that \texttt{gla1ve}'s distribution has a smaller standard deviation (0.0038) than \texttt{device} (0.0041) or \texttt{dupreeh} (0.0041). At the same time, can see that \texttt{device} exhibits a far higher average WPA per round, where as \texttt{dupreeh} and \texttt{gla1ve} are far more similar.

\subsection{Identifying High Impact Plays}
While WPA has clear uses for player scouting, tactical planning or betting, media and fans can also use win probability models to find high leverage or impactful plays. Information on these plays can be used for commentating or drafting articles and highlights. By assigning each action a value, we can easily query actions based on their change in win probability, or query actions based on the current win probability.

In Figure~\ref{fig:big-play}, we see an example of one of the highest win probability swings for a round in our data. The T player \texttt{ZywOo}, considered the best player by both our WPA framework and HLTV Rating 2.0, came from behind with just 13 health points to defeat the two remaining CT. The highest valued action in this sequence was a headshot on \texttt{B1NGO}, a player from \texttt{EHOME}, which provided a gain of 67\% in win probability. \texttt{ZywOo} then killed the remaining CT player to win the round.

Because the value function in Equation~\ref{eq:value} takes two arbitrary game states as arguments, we can apply this function on two non-sequential game states. In the context of finding impactful plays, we can use this property to assess the total impact of a series of actions. In Figure~\ref{fig:big-play}, \texttt{ZywOo} became the sole player left on the T side around 100 seconds into the round. At this point, the 1-vs-2 situation presents under a 3\% chance of the T side winning. With this in mind, \texttt{ZywOo} effectively turned an untenable game situation into a round win.

\subsection{Limitations and Future Work}
\label{discussion:limitations-future-work}
One of the limitations of the current WPA framework is that it only considers damage events. However, there are many events besides damages attributable to players such as bomb plants/defuses and movement. Oftentimes, these events can have unclear attributions. For example, if the T side plants the bomb and there are three CT remaining, it is not immediately obvious how the loss in win probability should be divided. A naive method could be to distribute the loss equally. On the other hand, it might not be accurate to detract from a player who had no part in losing control of the bombsite. In the future, we will consider a broader set of actions when calculating WPA. Another action type to consider are grenade throws. Grenades change the context of the game, in a non-damage point of view, by blinding opponents or obstructing views. It is clear that killing an opponent who is blinded is easier than an opponent who is not. Additionally, many kills through smoke grenades often occur through players randomly attacking through the smoke. Our future work will acknowledge how grenades change the context of the game.

In future work, we want to fully explore spatially derived features using our graph distance measure. For example, distances between players in a team or between their opponents may reveal tactics that influence a team's win probability. We plan to use this spatial information to automatically classify certain movements, plays or tactics. We also intend to explore other distance measures. One of the downsides of our graph distance metric is that tile size in the navigation mesh is not uniform. Therefore, using a metric such as geodesic distance could further improve our performance.

It is important to note that our framework does not consider the level of the match being played. Thus, a strong assumption of the WPA framework is that analyzed teams are of similar skill categories. Win probability based valuation models in other sports also tend to disregard the level of the teams involved in the match, as it is very hard to obtain reliable estimates of team performance. One method to explore in the future is using pre-match betting spread or rankings in the win probability model, to capture the skill differences between the teams. In this way, WPA could be used across all settings. Sports such as baseball, basketball and soccer have significant literature on developing team ranking systems that future work may consider, such as the works described in Lopez~et.~al.~\cite{lopez2018often}.
\section{Conclusion}
\label{sec:conclusion}
This paper introduces (1) a data model for CSGO designed to facilitate CSGO data analysis, (2) a graph based distance measure to describe distances in esports and (3) WPA, a context-aware framework to value players based on their actions. 
% HD: Removed this since we are supposed to still not have made it public
%Additionally, we provide an open-source library that facilitates CSGO data analysis using our data model. 
We find that WPA provides a reproducible, consistent and unique dimension by which teams, fans, media and gamblers can assess the skill and variability of CSGO players at all levels of the sport. 

\section*{Acknowledgments}

This work was partially supported by NSF awards: CNS-1229185, CCF-1533564, CNS-1544753, CNS-1626098, CNS-1730396,  CNS-1828576; and the NYU Moore Sloan Data Science Environment. 

\balance

\bibliographystyle{IEEEtran}  
\bibliography{csgo}  

\end{document}